\documentclass[10pt,twocolumn,letterpaper]{article}

\usepackage{cvpr}              

\usepackage{wrapfig}
\usepackage{multirow}
\usepackage{stfloats}
\usepackage{makecell}
\usepackage{float}
\usepackage{lipsum}
\usepackage{caption}

\definecolor{cvprblue}{rgb}{0.21,0.49,0.74}
\usepackage[pagebackref,breaklinks,colorlinks,allcolors=cvprblue]{hyperref}

\def\paperID{****} 
\def\confName{CVPR}
\def\confYear{2026}

\title{MultiAnimate: Pose-Guided Image Animation Made Extensible}

\author{
    Yingcheng Hu\textsuperscript{1,2,3*} \quad
    Haowen Gong\textsuperscript{3} \quad
    Chuanguang Yang\textsuperscript{1}  \quad
    Zhulin An\textsuperscript{1,\dag} \\
    Yongjun Xu\textsuperscript{1} \quad
    Songhua Liu\textsuperscript{3,\dag} \\
    \textsuperscript{1}State Key Laboratory of AI Safety, Institute of Computing Technology, \\ Chinese Academy of Sciences \quad
    \textsuperscript{2}ShanghaiTech University \quad
    \textsuperscript{3}Shanghai Jiao Tong University
    \\
    \small \textsuperscript{\dag}Corresponding Authors
}

\newcommand\blfootnote[1]{%
  \begingroup
  \renewcommand\thefootnote{}\footnote{#1}%
  \addtocounter{footnote}{-1}%
  \endgroup
}

\begin{document}

\twocolumn[{%
\vspace{-1.5cm}
\maketitle
\vspace{-1cm}
\begin{figure}[H]
  \hsize=\textwidth 
  \centering
  
  \begin{minipage}{\textwidth} 
    \centering
    \includegraphics[width=\textwidth]{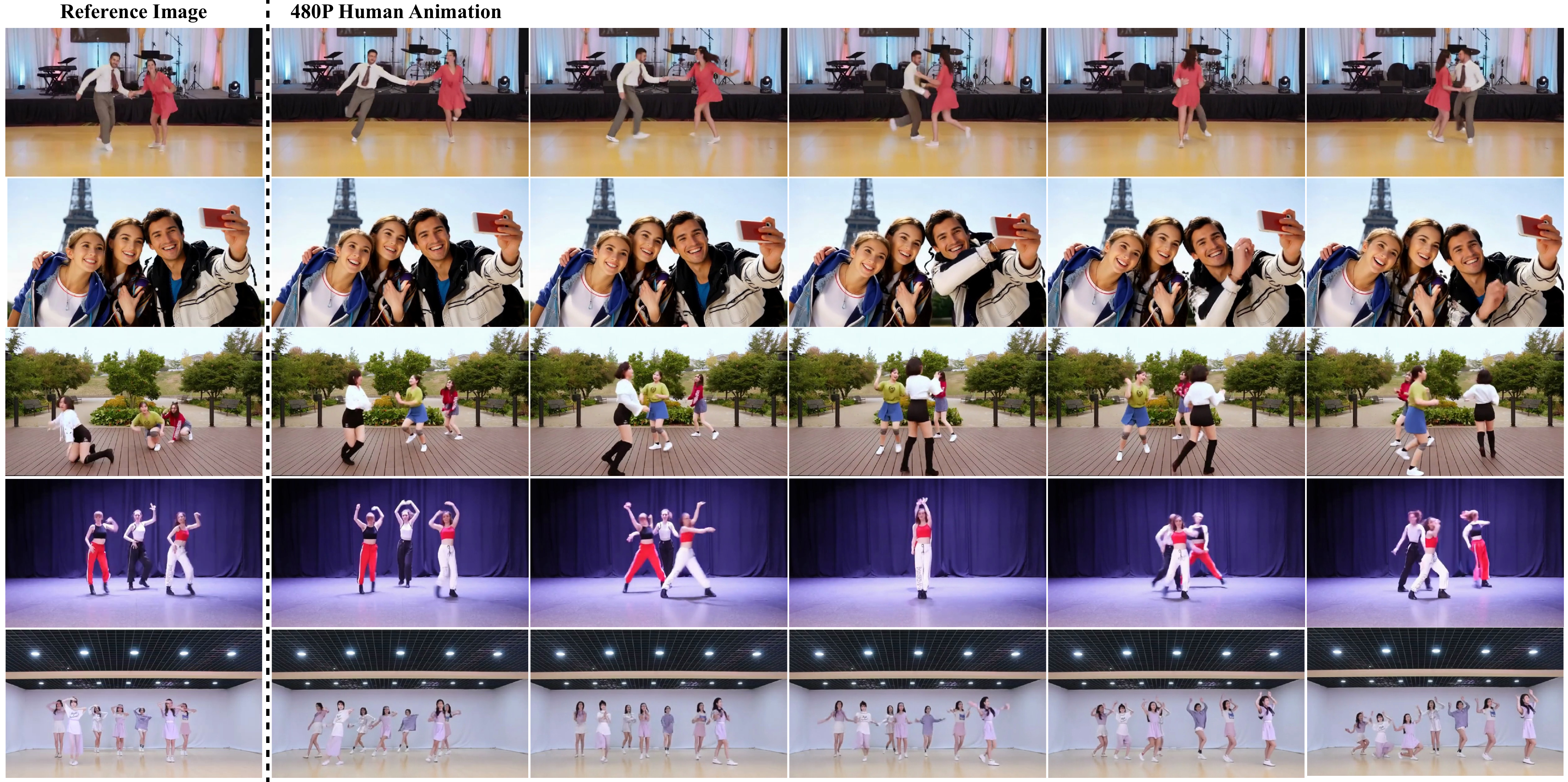} 
    \vspace{-0.5cm}
    \caption{
      \textbf{Multi-character pose-guided image animation generated by our framework.}
      Our method performs multi-character image animation with consistent identity and appearance for each character.
      Notably, our framework, trained only on two-character data, is capable of producing identity-consistent three-person videos and can, in principle, be extended to scenarios with even more participants (e.g., seven characters).
    }
  \end{minipage}
  
\end{figure}
}]
\blfootnote{*Work done during a research internship at the School of Artificial Intelligence, Shanghai Jiao Tong University.
}

\begin{abstract}

Pose-guided human image animation aims to synthesize realistic videos of a reference character driven by a sequence of poses. 
While diffusion-based methods have achieved remarkable success, most existing approaches are limited to single-character animation. 
We observe that naively extending these methods to multi-character scenarios often leads to identity confusion and implausible occlusions between characters. 
To address these challenges, in this paper, we propose an extensible multi-character image animation framework built upon modern Diffusion Transformers (DiTs) for video generation. 
At its core, our framework introduces two novel components-Identifier Assigner and Identifier Adapter—which collaboratively capture per-person positional cues and inter-person spatial relationships. 
This mask-driven scheme, along with a scalable training strategy, not only enhances flexibility but also enables generalization to scenarios with more characters than those seen during training. 
Remarkably, trained on only a two-character dataset, our model generalizes to multi-character animation while maintaining compatibility with single-character cases. 
Extensive experiments demonstrate that our approach achieves state-of-the-art performance in multi-character image animation, surpassing existing diffusion-based baselines. 
Video demos are available at
\url{https://hyc001.github.io/MultiAnimate/}.
\end{abstract}

\section{Introduction}
\begin{figure}[!h]
  \centering
  \begin{subfigure}[b]{1\linewidth}
    \centering
    \includegraphics[width=\linewidth]{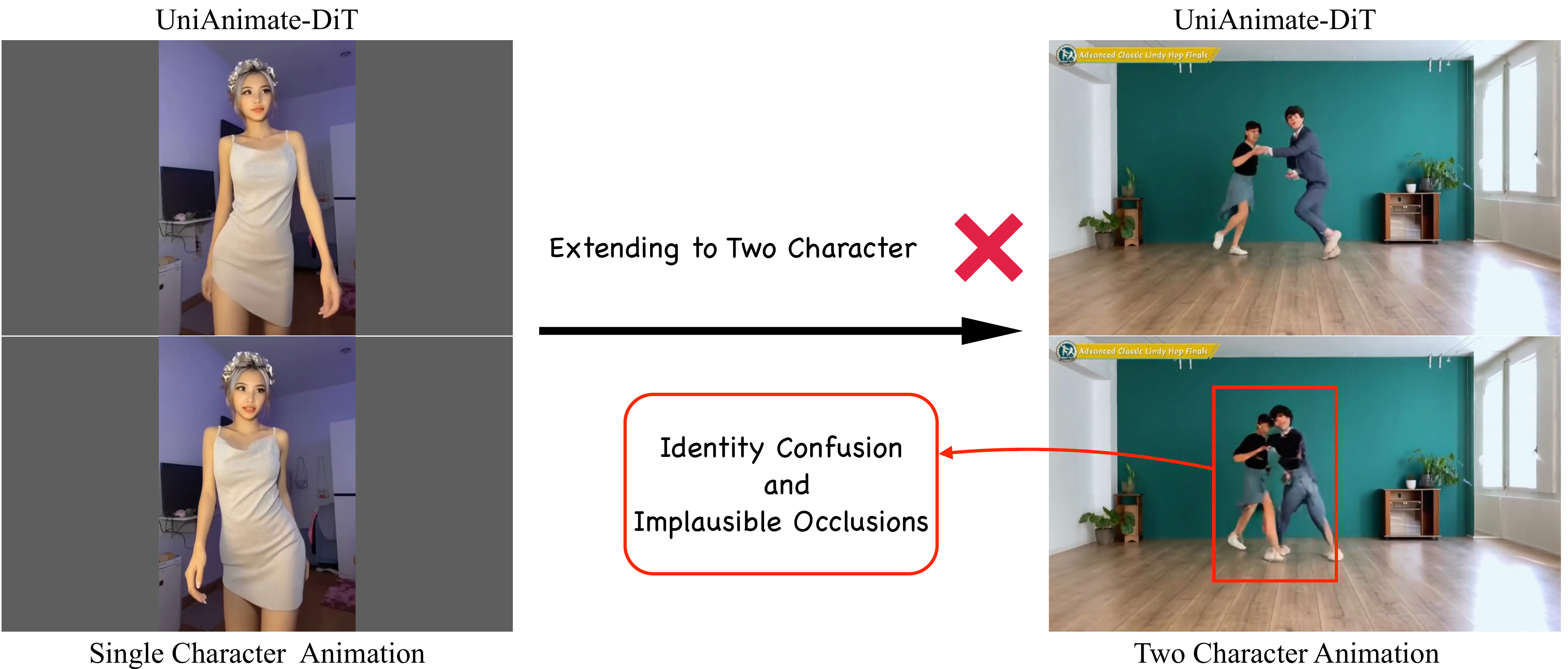}
    \vspace{-0.4cm}
    \caption{UniAnimate-DiT fails to extend to two character animation.}
    \label{motivation1}
  \end{subfigure}
  
  \begin{subfigure}[b]{1\linewidth}
    \centering
    \includegraphics[width=\linewidth]{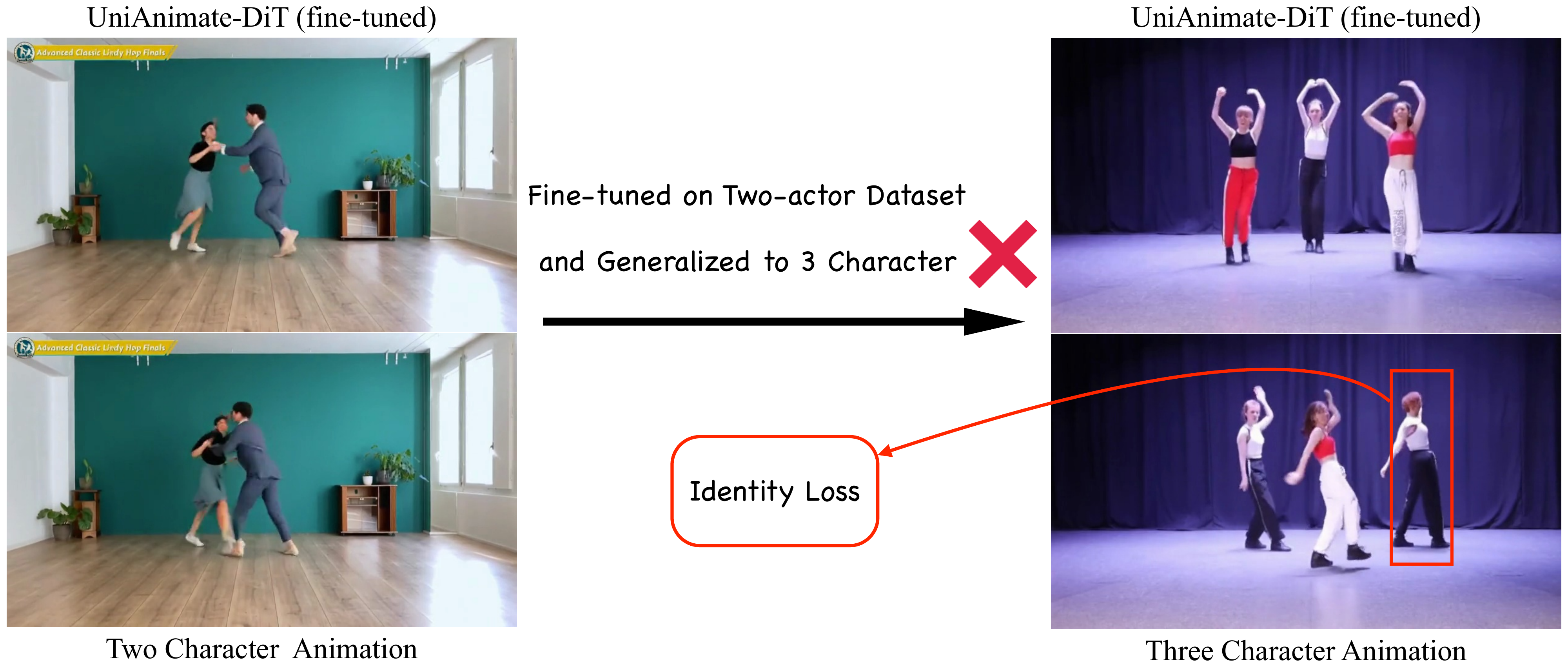}
    \caption{Fine-tuned UniAnimate-DiT fails to generalize to cases with unseen numbers of characters.}
    \label{motivation2}
  \end{subfigure}
  \vspace{-0.8cm}
  \caption{Dilemmas of current methods in multi-character image animation.}
  \label{motivation}
  \vspace{-0.6cm}
\end{figure}

Diffusion-based video generation \cite{blattmann2023stable,guo2023animatediff,kong2024hunyuanvideo,ho2022video,wan2025} has witnessed remarkable advances in recent years, providing powerful tools for creating dynamic visual content. 
Within this trend, controllable video generation \cite{li2024dispose,zhang2023adding,zhang2025mimicmotion,wang2025unianimate,yang2025multipartycollaborativeattentioncontrol} has emerged as a major research focus, as users increasingly seek to create visual content with customized layouts, motions, or identities.
A representative task is human image animation, which aims to synthesize a video of a reference character driven by a sequence of poses. 
This task has attracted significant interest due to its widespread applications, such as film production \cite{wu2025moviebench} and digital avatars \cite{xu2024mambatalk}.

Human image animation generally imposes two core requirements: (i) maintaining the consistent identity of the generated person with the given reference character and (ii) aligning the generated motion with the input pose sequence. 
Various works \cite{li2024dispose,zhang2025mimicmotion,wang2025unianimate,vace} have been proposed to address these issues and have made impressive progress. 
However, most existing methods primarily focus on single-character animation, while its natural extension-multi-character animation—remains underexplored. 

In fact, although this extension seems conceptually straightforward, it poses significantly greater challenges than it appears. 
On the one hand, it requires preserving per-character consistency while faithfully modeling the spatial interactions between multiple characters. 
However, as shown in \cref{motivation1}, we observe that naively extending single-character animation methods to multi-character scenarios often fails to meet these requirements, leading to identity confusion and implausible occlusions. 
On the other hand, scalability remains a major challenge. Although some prior works consider multi-character scenarios, existing human image animation models are typically trained with a fixed number of participants, \textit{e.g.}, two-person animation trained on two-person datasets. As illustrated in \cref{motivation2}, such models struggle to generalize to cases with unseen numbers of characters. Extending them to more participants requires collecting corresponding multi-person data and retraining or adapting the models accordingly—a process that is highly cumbersome if not infeasible at all. 

These observations raise a key question: \emph{Can we design a human image animation framework that is both robust and generalizable?} 
To this end, in this paper, we propose \textbf{MultiAnimate}, a novel multi-actor image animation framework that (1) preserves multi-character consistency and visual fidelity under complex interactions, and (2) exhibits strong scalability to scenarios with more characters than those seen during training. 

At the core of our approach lies an effective mechanism that resolves the inherently underdetermined association between poses and identities during spatial interactions, as illustrated in Fig.~\ref{rotate}. 
Specifically, we introduce two key modules-Identifier Assigner and Identifier Adapter-which assign a unique identifier to each character across all frames and embed this identity cue into the feature space of the Diffusion Transformer (DiT) backbone. 
This explicit identity encoding enables consistent character tracking throughout the video, effectively mitigating confusion in complex multi-character interactions.

While the identifier-based strategy effectively resolves character confusion, a remaining challenge lies in enabling these identifiers to generalize to unseen numbers of characters. 
To address this problem, we devise a scalable training strategy that randomly samples identifiers from a learnable embedding space at each iteration. 
By the time training converges, all identifiers have been explicitly trained to be mutually distinguishable. 
As a result, during inference, even when more characters, along with more identifiers, are introduced than were present during training, the model can naturally differentiate them and maintain consistent character identities. 
In our experiments, we demonstrate that training solely on two-character data is sufficient to perform multi-character image animation while maintaining compatibility with single-character cases. 

Moreover, to further enhance the visual quality of generated videos, we leverage modern video generation models, such as Wan 2.2~\cite{wan2025}, to synthesize high-quality multi-character training data as an additional supplement. 
Extensive experiments demonstrate that our approach surpasses existing diffusion-based methods in multi-character image animation while preserving competitive performance in single-character scenarios. 
In summary, our contributions are threefold:
\begin{itemize}
    \item Overall, we propose \textbf{MultiAnimate} for multi-character image animation, which is the first extensible framework built upon modern DiT-based video generators, to the best of our knowledge. 
    \item Technically, we (i) introduce two novel modules-Identifier Assigner and Identifier Adapter—to jointly and explicitly model per-person spatial features and inter-person interactions, and (ii) propose a scalable training strategy that allows our model to handle multi-character image animation with more characters than those present in the training data.
    \item Empirically, we (i) curate a new high-quality dataset for our experiments to overcome the limitations of prior training data in frame quality, and (ii) perform experiments demonstrating that our framework achieves superior performance in multi-character video generation while remaining competitive in single-character scenarios.
\end{itemize}

\section{Related Works}
\noindent\textbf{Diffusion models for video generation.} Diffusion models \cite{dhariwal2021diffusion,ho2020denoising,ho2022cascaded,nichol2021improved,rombach2022high,song2020denoising,song2020score,chen2025dparallel,wang2026sparsed,feng2024relational} have rapidly advanced video generation \cite{ho2022video,poole2022dreamfusion} in recent years. 
Many approaches build upon UNet \cite{ronneberger2015u}, adding temporal modules to image diffusion models \cite{ho2022classifier,ramesh2022hierarchical, saharia2022photorealistic} for spatio-temporal modeling \cite{blattmann2023stable,guo2023animatediff}. 
More recently, inspired by the success of DiT \cite{peebles2023scalable} in image generation, some works \cite{chen2024gentron,ma2024latte,lin2024open,feng2025q} replace the UNet with DiT blocks to further improve generation quality. 
3D VAE architectures are introduced in Hunyuan-Video \cite{kong2024hunyuanvideo} and Wan-Video \cite{wan2025} to enhance video compression and reconstruction. 
These DiT-based frameworks flexibly inject conditions via cross-attention or feature concatenation, and have become increasingly popular foundations for controllable video generation upon which our work is built.

\noindent\textbf{Pose-guided human image animation.} 
Pose-guided human image animation typically uses skeleton poses \cite{zhang2024evaluationagent,cao2017realtime} to drive the motion of a target character. 
Current works introduce additional conditioning signals to improve controllability and visual quality. 
Champ \cite{zhu2024champ} and VidvidPose \cite{wang2024vividpose} incorporate depth and 3D body priors such as SMPL to enhance controllable motion synthesis. StableAnimator \cite{tu2025stableanimator} introduces a dedicated face encoder combined with face masks to better preserve facial identity. 
However, these approaches mainly focus on single-character animation. 
Recently, Follow-Your-Pose v2 \cite{xue2024follow} leverages optical flow and depth cues to better handle occlusions between interacting people. 
DanceTogether \cite{chen2025dancetogether} jointly exploits mask and pose features to model spatial relationships between various characters. 
Unlike them, we target a scalable multi-character image animation framework that can generalize to scenarios with more characters than those seen in training.

\section{Methods}
For pose-guided multi-character image animation, given a reference image $I_{\text{ref}} \in \mathbb{R}^{3 \times H \times W},$ 
a driving pose sequence $P \in \mathbb{R}^{T \times 3 \times H \times W},$ and a set of per-person tracking masks $\{M_i\}_{i=1}^{n}, $ $M_i \in \mathbb{R}^{T \times 1 \times H \times W},$ where $T$ denotes the number of frames and $n$ is the number of tracked persons, our goal is to generate a target video $V_{\text{tar}} \in \mathbb{R}^{T \times 3 \times H \times W},$ that (1) preserves each identity in $I_{\text{ref}}$, (2) aligns with the motion in pose sequence $P$, and (3) aligns with the spatial relationships among per-person tracking masks $\{M_i\}_{i=1}^{n}$. 

In Sec.~\ref{sec:3-1}, we present the design of our pipeline, which models per-person positional cues and inter-person spatial relationships through the proposed Identifier Assigner and Identifier Adapter. 
In Sec.~\ref{sec:3-2}, we describe our training strategy that enables multi-character image animation with more characters than those seen during training.

\subsection{MultiAnimate}\label{sec:3-1}
In this section, we first analyze the problem in pose-guided multi-character image animation and present our pipeline.
\begin{figure}[tb]
  \centering
   \includegraphics[width=1\linewidth]{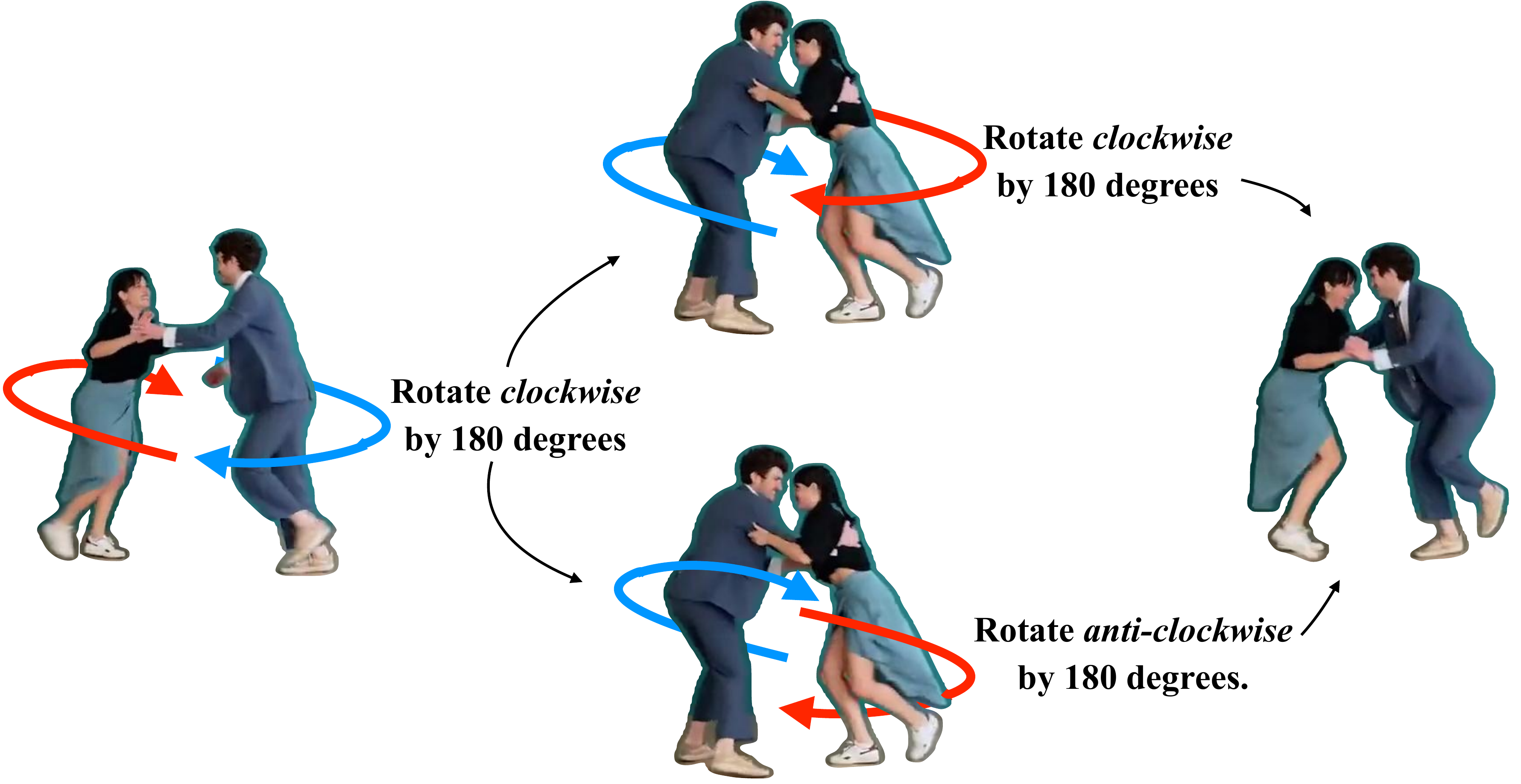}
   \caption{
In multi-character image animation, identical pose sequences can lead to multiple plausible motion trajectories.
}
   \label{rotate}
\vspace{-0.4cm}
\end{figure}

\subsubsection{A Multi-Character Motion Ambiguity} 
Consider simply extending single-character to multi-character pose-guided image animation. 
In single-character setting, the mapping between the reference image and the pose sequence is unique. 
However, when multiple persons are generated simultaneously, an ambiguity arises. 
As illustrated in \cref{rotate}, the two people first rotate $180^{\circ}$ clockwise and exchange their positions. 
From this intermediate state, there exist two visually plausible continuations: (1) both characters continue rotating clockwise by another $180^{\circ}$ to return to their initial positions, or (2) both characters rotate counterclockwise by $180^{\circ}$ back to the same configuration. 
These two trajectories can be yielded from nearly identical pose sequences, making the output non-unique unless additional spatial cues are provided. 
Therefore, in multi-character pose-guided image animation, per-person tracking masks are necessary to fix a unique correspondence between each character and its motion trajectory.

A straightforward way to incorporate these cues is to extract pose and tracking mask features for each person individually, bind them together, and then aggregate all per-person representations by summation, as done in concurrent work \cite{chen2025dancetogether}. 
However, such a design requires additional data processing to extract a pose sequence for each character, 
and direct summation cannot ensure that spatial relationships between different people are adequately preserved. 

\subsubsection{Proposed Pipeline}
\begin{figure*}[!t]
  \centering
   \includegraphics[width=1\linewidth]{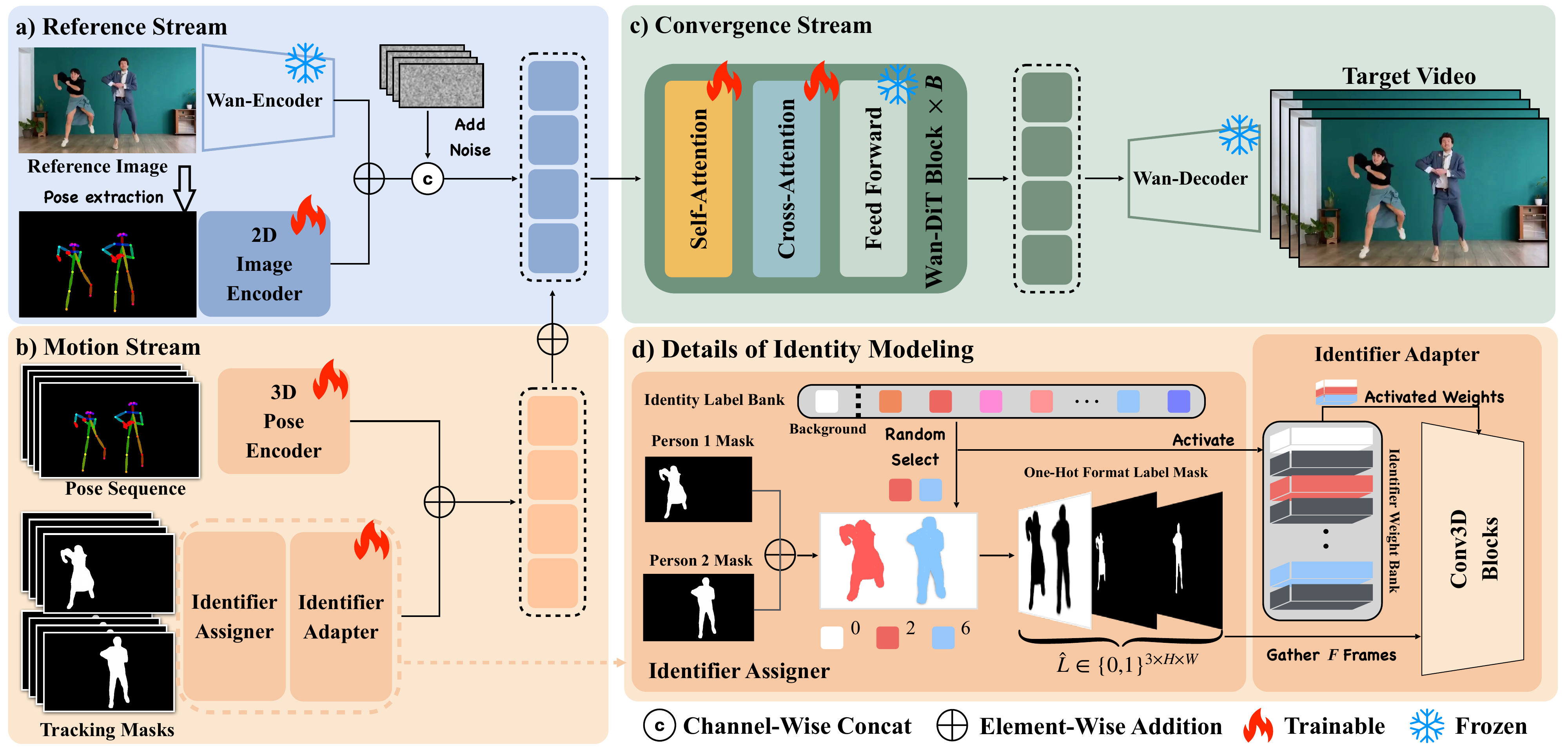}
   \vspace{-0.7cm}
   \caption{
\textbf{Overview of our framework.}
Our pipeline contains two main streams: 
the \textit{reference stream}, which encodes the reference image and its pose to capture appearance information, 
and the \textit{motion stream}, which encodes multi-character pose sequences and tracking masks to model motion and spatial conditions. 
The two streams are fused through element-wise addition of latent tokens. 
The Identifier Assigner unifies per-person tracking masks into a structured label representation, 
preserving spatial relationships and interactions among multiple characters. 
This representation is converted to the feature space of the DiT backbone by the Identifier Adapter.
}
   \label{pipeline}
\vspace{-0.4cm}
\end{figure*}

\noindent\textbf{Data flow.}  
Following previous work \cite{wang2025unianimate}, we build our framework upon the Wan2.1 \cite{wan2025} I2V architecture. 
As illustrated in \cref{pipeline}, our pipeline consists of two main processing streams: one for encoding the reference appearance and the other for encoding the motion and spatial conditions. 

\begin{itemize}[leftmargin=1.0em]
    \item \textbf{Reference stream.}  
    The reference image $I_{\text{ref}} \in \mathbb{R}^{3 \times H \times W}$ is first passed through a VAE Encoder to obtain its latent representation.  
    The reference pose extracted from $I_{\text{ref}}$ is processed by an Image Encoder composed of stacked 2D convolutional layers. 
    Their outputs are added, concatenated with random noise, and further patchified to serve as the input tokens.

    \item \textbf{Motion stream.}  
    The pose sequence $P \in \mathbb{R}^{T \times 3 \times H \times W}$ is encoded by a 3D convolutional Pose Encoder to capture temporal motion dynamics.  
    Meanwhile, the per-person tracking masks $\{M_i\}_{i=1}^{n}$ are first processed by the Identifier Assigner to unify and encode all masks into a structured label representation. 
    The resulting feature maps are then fed into the Identifier Adapter, which models per-person positional features and inter-person spatial relationships.
    The encoded pose and mask features are first added and then fused with patchified latent tokens from the reference stream through element-wise addition. 
\end{itemize}

\noindent\textbf{Identifier Assigner.}  
Given the set of per-person tracking masks $\{M_i\}_{i=1}^{n}$, we first perform a preprocessing step, referred to as the Identifier Assigner in \cref{pipeline}. 
This module unifies all per-person masks into a single label map $\mathcal{L} \in \mathbb{R}^{H \times W}$, where background pixels are labeled as $0$. 
Consider the case where only a two-character dataset is available, 
For the two characters A and B, all pixels in their tracking masks with values greater than $0$ are assigned labels $a$ and $b$ in $\mathcal{L}$, respectively, where $a$ and $b$ are two distinct non-zero identifiers from Identity Label Bank. 
As a result, we obtain a label map with values in $\{0, a, b\}$, denoted as $\mathcal{L} \in \{0, a, b\}^{H \times W}$.  
We then apply one-hot encoding to $\mathcal{L}$, producing a binary tensor
$\hat{\mathcal{L}} \in \{0, 1\}^{3 \times H \times W},$
where the three channels correspond to the background and the two characters, each encoding its spatial occupancy. 

The Identifier Adapter, consisting of stacked 3D convolutional layers, then takes $\hat{\mathcal{L}}$ as input. 
The design of the Identifier Adapter, together with the associated training strategy, will be detailed in the following subsection.
Our design preserves spatial relationships among multiple characters and captures interpersonal interactions, such as proximity and occlusion. 
This provides an explicit spatial prior for generating videos of multiple people.

\subsection{Training Strategy}\label{sec:3-2}
In this section, we first discuss the challenges encountered during training and then present our strategy.

\subsubsection{Symmetry Issue and Generalization Ability}
\begin{figure}[tb]
  \centering
   \includegraphics[width=1\linewidth]{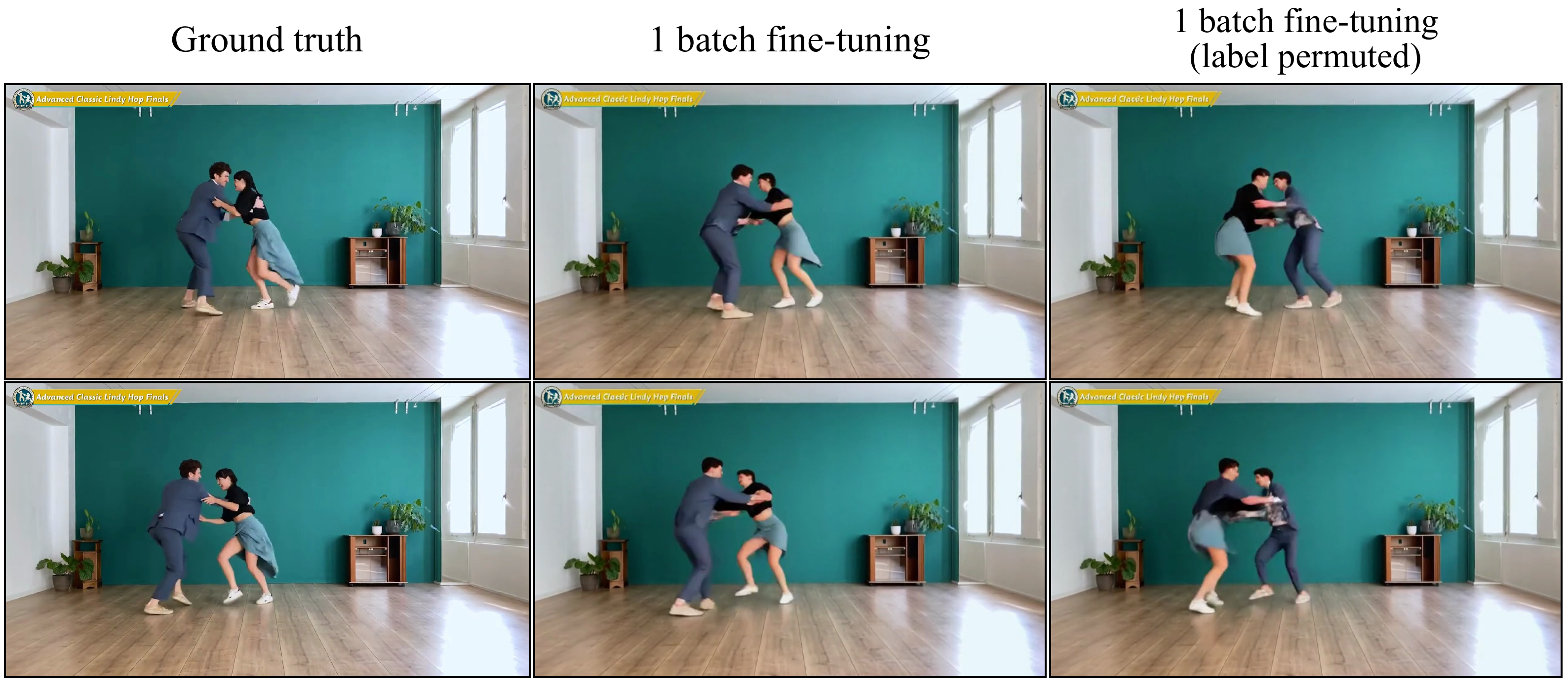}
\vspace{-0.5cm}
   \caption{
Our framework performs well at early training stages, 
but inconsistencies emerge when the person-assigned labels at inference differ from those seen during training. 
}
   \label{label_permuted}
\vspace{-0.4cm}
\end{figure}

Our pipeline can achieve promising results at early stage of training, as shown in \cref{label_permuted}. 
However, a \textit{symmetry problem} arises during training. 
Specifically, if the person-assigned labels during inference differ from those used in training, the generated results become inconsistent. 
This suggests that during training, each person's position is associated with a particular channel in the Identifier Adapter, as opposed to being linked to its corresponding tracking mask. 
Such channel-identity binding contradicts our design goal, which is to let the model associate each person with its tracking mask, not with a fixed label or channel.  

Additionally, current pose-guided human image animation frameworks are typically trained with a fixed number of participants, \textit{e.g.}, two-person animation trained on two-person datasets.
As collecting multi-person data is costly, 
such a setup severely limits generalization of existing frameworks to more complex multi-character scenarios. 
We target a flexible framework that generalizes to a varying number of participants, even with more characters than those seen during training.

\subsubsection{Proposed Strategy}
Due to our mask-driven design, the above two goals can be effectively achieved.
Consider that the model is expected to support up to $n$-character image animation during inference. 
In the first Conv3D layer of the Identifier Adapter, we introduce an Identifier Weight Bank storing $n$ identifier channels. 
Under a two-character training setting, only three channels, including identifiers for the two characters and the background, are activated in each training iteration.

Specifically, in each training iteration, we randomly assign identity labels to the two characters from the Identity Label Bank of size $n$. 
These assigned labels are also used to activate the corresponding weights in the Identifier Weight Bank. 
By the time training converges, although the training dataset contains only two-character videos, all the $n$ channels in the Identity Weight Bank are seen and trained to be mutually distinct. 
In this way, our framework can robustly generate videos with up to $n$ characters during inference. 
Moreover, since labels are randomly assigned, the network learns to associate each character with its spatial mask rather than a fixed channel, which helps alleviate the symmetry problem. 
\begin{table*}[t]
    \centering
    \vspace{-0.2cm}
    \caption{
Quantitative comparison with other SOTA methods on the test split of Swing Dance,Gen-dataset and Unseen dance videos.
}
    \small
    \begin{tabular}{c|c|cccc|cc}
        \hline
        
\multirow{1}{*}{\textbf{Dataset}} &
\multirow{1}{*}{\textbf{Method}} &
\multirow{1}{*}{\textbf{PSNR}$\uparrow$} &
\multirow{1}{*}{\textbf{SSIM}$\uparrow$} &
\multirow{1}{*}{\textbf{L1(E-3)}$\downarrow$} &
\multirow{1}{*}{\textbf{LPIPS}$\downarrow$} &
\multirow{1}{*}{\textbf{FVD}$\downarrow$} &
\multirow{1}{*}{\textbf{FID-VID}$\downarrow$} \\

        \hline

\multirow{5}{*}[-0.1cm]{\textbf{Swing Dance}}
        &
        Mimicmotion \cite{zhang2025mimicmotion} & 13.47 & 0.390 & 145.10 & 0.501  & 780.31 & 49.69\\

        &
        DisPose \cite{li2024dispose} & 14.08 & 0.444 & 137.69 & 0.466  & \underline{746.29} & 46.24\\

        &
        UniAnimate-DiT \cite{wang2025unianimate} & \underline{16.15} & \underline{0.619} &\underline{98.72} & \underline{0.427} &891.89  &\underline{27.71}\\

        &
        VACE \cite{vace}  & 11.15 & 0.311 & 216.98 & 0.563 &763.75 & 29.88\\


        &
        Ours (\textit{Stage 1 model})&\textbf{19.40} & \textbf{0.687} & \textbf{64.94} & \textbf{0.335} & \textbf{648.84} & \textbf{22.50}\\
    \hline
\multirow{5}{*}[-0.1cm]{\textbf{Gen-dataset}}
        &
        Mimicmotion \cite{zhang2025mimicmotion} & 11.96 & 0.286 &173.14  & 0.444  & 865.36 & 86.99\\

        &
        DisPose \cite{li2024dispose} & 15.33 & 0.488 & 105.41 & 0.266  & 712.61 & 53.70\\

        &
        UniAnimate-DiT \cite{wang2025unianimate} & \underline{18.10} &\underline{0.630} & \underline{80.52} & 0.262  & \underline{565.50} & \textbf{32.10}\\

        &
        VACE \cite{vace} & 16.39 &0.559  & 97.14 & \textbf{0.231}  & 657.23 & 38.86\\

        &
        Ours (\textit{Stage 1 model})& \textbf{19.05} &\textbf{0.659}  &\textbf{73.01}  & \underline{0.251}  & \textbf{454.92} & \underline{33.57}\\
    \hline
\multirow{5}{*}[-0.1cm]{\makecell{\textbf{Unseen}\\\textbf{dance videos}}}
        &
        Mimicmotion \cite{zhang2025mimicmotion} & 13.53 & 0.376 & 145.73 & 0.381  & 1634.12 & 145.67\\

        &
        DisPose \cite{li2024dispose} & 15.88 & 0.694 & 92.63 & 0.304  & 1133.57 & 97.23\\

        &
        UniAnimate-DiT \cite{wang2025unianimate} & \underline{17.94} & \underline{0.751} & \underline{70.07} & 0.286  & \underline{624.45} & \underline{71.24}\\

        &
        VACE \cite{vace} & 17.24 & 0.714 & 84.34 & \underline{0.279}  & 922.66 & 78.93\\

        &
        Ours (\textit{Extended model}) & \textbf{23.24} & \textbf{0.857} & \textbf{33.40} & \textbf{0.185}  & \textbf{358.74} & \textbf{43.12}\\
    \hline
    \end{tabular}
    \label{tab:main}
    \vspace{-0.2cm}
\end{table*}

\section{Experiments}
\noindent\textbf{Datasets.}  
We used the Swing Dance dataset \cite{maluleke2024synergy}, which comprises approximately 30 hours of swing dance videos featuring 680 unique couple subjects. This dataset is utilized to train the model to learn complex interactions and spatial relationships between individuals.
Also, we generated 2079 five-second videos using the Wan 2.2 Video Generator~\cite{wan2025}, denoted as the \textit{Gen-dataset}. 
It contains various scenes with two or three characters and is primarily used to assess the adaptability of the model to varying scenarios. 
Optionally, it can also be incorporated into training to further enhance robustness of the model.
In addition, the TiKToK dataset \cite{Jafarian_2021_CVPR_TikTok}, along with over 50 minutes of unseen dance videos featuring 3-7 characters from social media,
serves solely to evaluate the generalization capability of our framework.

\noindent\textbf{Evaluation Metrics and Baselines.}  
The quality of generated videos is evaluated using FVD \cite{unterthiner2018towards} and FID-VID \cite{balaji2019conditional}. To further evaluate the frame generation quality, we report frame-wise PSNR \cite{hore2010image}, SSIM \cite{wang2004image}, L1 loss and LISPIS \cite{zhang2018unreasonable}.
We compare our method with other open-source pose-guided human image animation models, including UniAnimate-DiT \cite{wang2025unianimate}, VACE \cite{vace}, MimicMotion \cite{zhang2025mimicmotion}, and DisPose \cite{li2024dispose}.

\noindent\textbf{Implementation Details.}  
We utilize DWPose \cite{yang2023effective} and Sa2VA \cite{sa2va} to extract skeleton poses and tracking masks. We initialized the Image Encoder, Pose Encoder and Lora with pre-trained weights in UniAnimate-DiT \cite{wang2025unianimate}. Our pipeline is first trained on the training split of the Swing Dance dataset for 40 epochs, 7000 training steps, and supports up to 3-character animation, referred to as the \textit{Stage 1 model}. Followed by 3 epochs, totally 2400 steps training on the training split of the Gen-dataset, referred to as the \textit{Stage 2 model}. Additionally, we extended the \textit{Stage 1 model} to support up to 7-character image animation. This version was trained on the training split of the Swing Dance dataset for 24 epochs (4,200 steps) and is referred to as the \textit{Extended model}. The training is performed on 2 NVIDIA A100 80GB GPUs with a batch size of 1 per GPU and the learning rate is set to $1 \times 10^{-4}$.


\begin{figure*}[tb]
  \centering
   \includegraphics[width=1\linewidth]{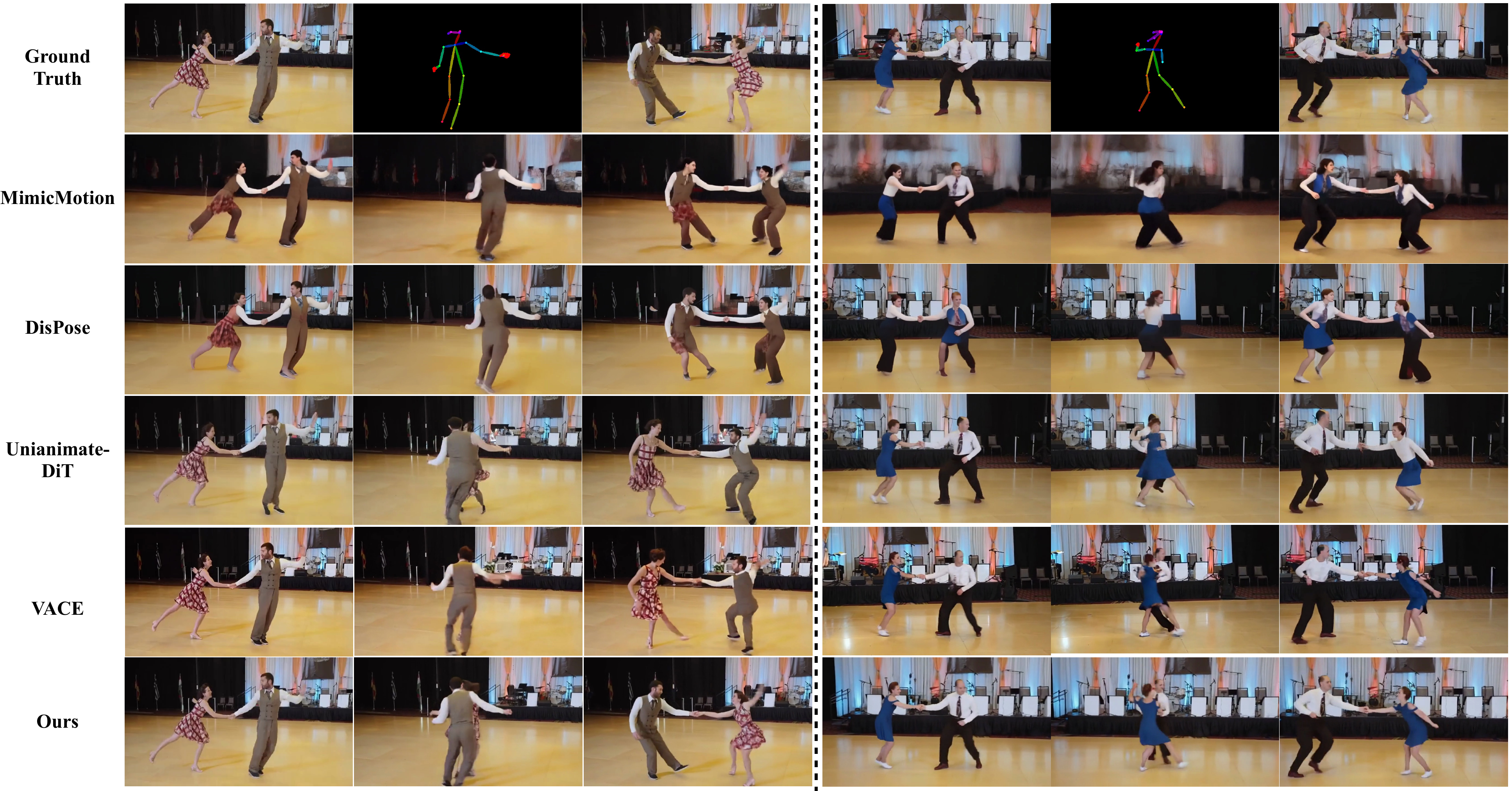}
   \vspace{-0.2cm}
   \caption{
\textbf{Quality comparison with state-of-the-art methods on the Swing Dance dataset.}
More examples are provided in the supplementary material.
}
   \label{main}
\vspace{-0.4cm}
\end{figure*}

\subsection{Comparison with Concurrent Work}
\noindent\textbf{Quantitative Results.}  
We evaluate \textit{Stage 1 model} and \textit{Extended model}, which are trained only on the training split of Swing Dance dataset. 
As shown in \cref{tab:main}, we compare our method with state-of-the-art approaches on the test splits of the Swing Dance dataset, the Gen-dataset, and on unseen dance videos.

On the Swing Dance dataset, which emphasizes complex motion interactions between two characters, our method consistently achieves better scores than competing methods for all metrics. 
The advantage in FVD and FID-VID indicates that our approach produces temporally coherent and perceptually plausible videos.
Other evaluation metrics suggest that our model produce high frame-wise image quality.

On the Gen-dataset, where motions are relatively simple but scenes are more diverse, our model-without any additional training on this dataset-still achieves strong performance. 
A slight disadvantage in LPIPS and FID-VID may suggest our method is less adapted to highly diverse environments.

On the unseen dance videos, where character interactions are more frequent and complex and the scenes are more challenging.
Nevertheless, our method maintains a clear advantage, demonstrating strong generalization ability from two-actor training data to more complex multi-actor scenarios.



\noindent\textbf{Qualitative Results.}  
We present several visual results from the Swing Dance dataset. 
As shown in \cref{main}, both DisPose and MimicMotion suffer identity confusion at the beginning of image animation, and the background appear blurry. 
Although UniAnimate-DiT and VACE maintain better identity consistency in the early stage, they fail to preserve it after complex inter-person interactions, leading to identity swaps and unnatural occlusions. 
In contrast, our method produces visually coherent results, maintaining consistent identities and clear spatial relationships throughout the video.

\subsection{Various Animation Scenarios}
\begin{figure*}[tb]
  \centering
   \includegraphics[width=1\linewidth]{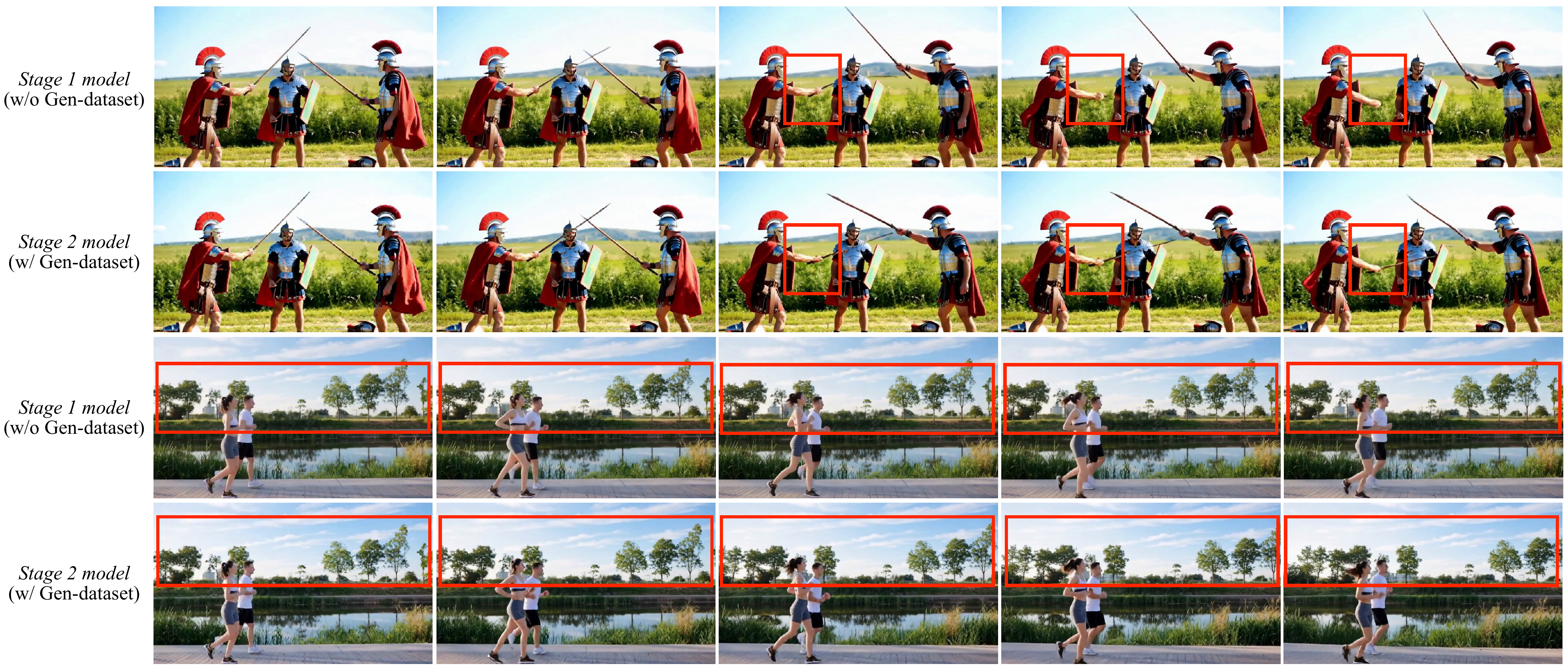}
   \caption{Training on the Gen-dataset enhances the model’s ability to maintain temporal consistency and adapt to diverse motion–scene interaction. }
   \label{ablation2}
   \vspace{-0.4cm}
\end{figure*}

To further enhance generalization, we additionally trained our model on the Gen-dataset. 
The examples shown in \cref{ablation2} are sampled from the Gen-dataset. As illustrated, the first example highlights (in the red box) the weapon held by the character in the generated video. Without training on the Gen-dataset, the \textit{Stage 1 model} fails to maintain subject consistency: the weapon disappears during motion generation. In contrast, the \textit{Stage 2 model}, preserves the weapon throughout the video, demonstrating improved temporal consistency.  
In the second example, the red box highlights the background region. The \textit{Stage 1 model} produces a nearly static background, showing limited scene dynamics. While the \textit{Stage 2 model} generates a background that changes naturally with the character's movement, making the camera view transitions more realistic. 
In general, training Gen-dataset enhances the model’s ability to capture motion-scene interactions in multi-person video generation.

\subsection{Ablation Studies}
\begin{figure}[t]
  \centering
   \includegraphics[width=1\linewidth]{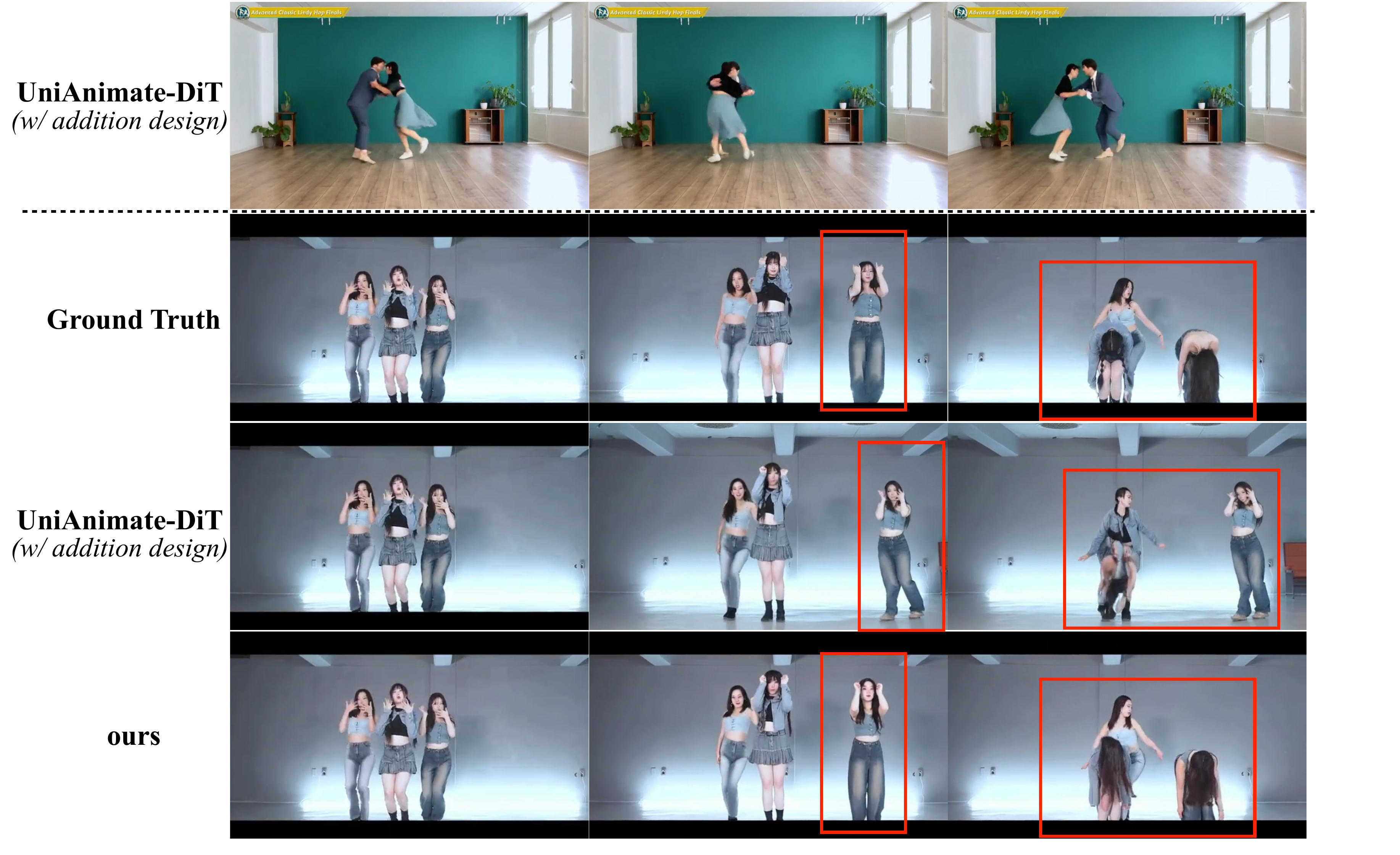}
   \caption{Addition-driven design works for two characters but fails to generalize to more characters.}
   \label{addtion}
\vspace{-0.4cm}
\end{figure}

\noindent\textbf{Mask-driven Design vs. Addition-driven Design.}
An alternative to multi-characters image animation is to aggregate per-characters representations through a simple summation strategy. 
Based on UniAnimate-DiT, we first isolate per-person poses using tracking masks, extract character-specific features for each individual, and then sum these features into a single combined representation.
This approach largely avoids the \textit{symmetry issue}. 
As shown in \cref{addtion}, with sufficient training on a two-characters dataset, the addition-driven design can handle two-character animation; 
However, it fails to generalize to settings with more characterss than those seen during training. 
In contrast, our mask-driven design preserves clear per-characters spatial organization and achieves much stronger scalability in multi-characters scenarios.

\begin{figure*}[tb]
  \centering
   \includegraphics[width=1\linewidth]{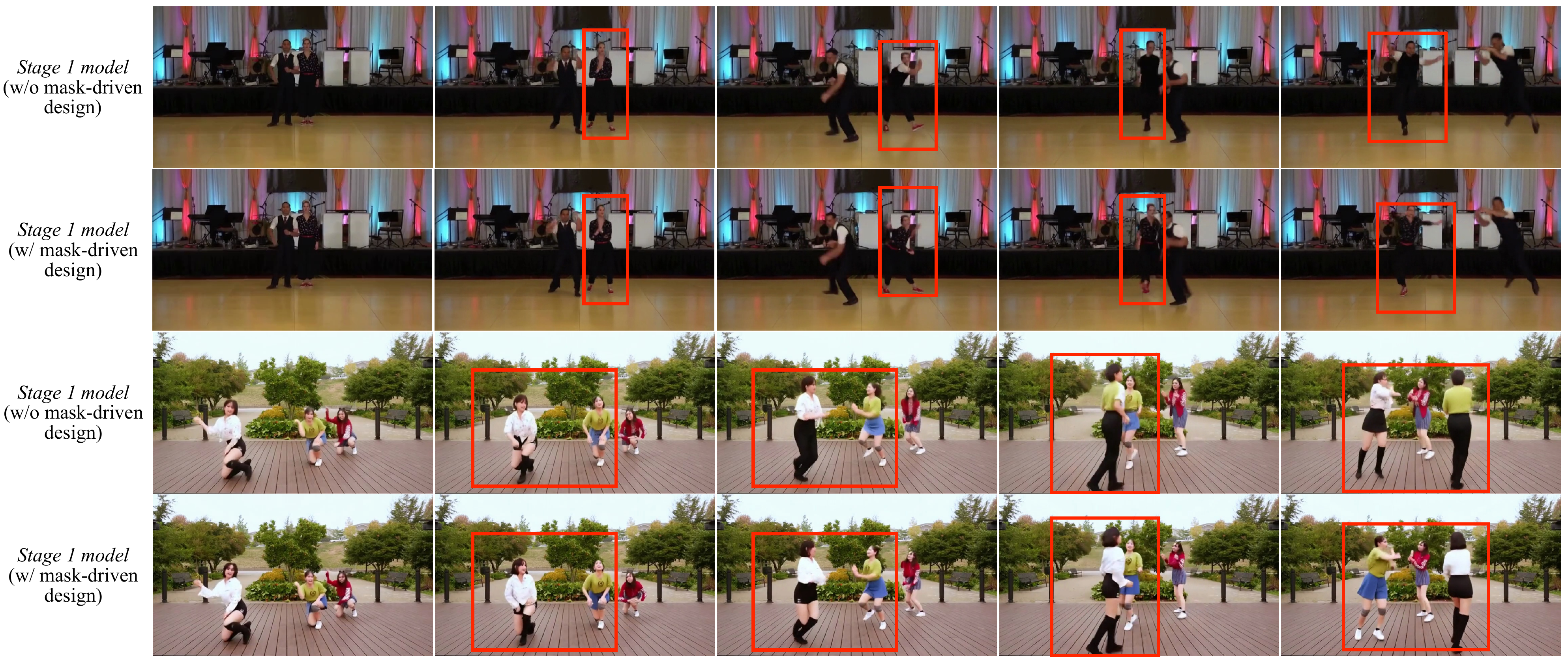}
   \caption{\textbf{Ablation on Identifier Assigner \& Identifier Adapter.} 
The combination of Identifier Assigner and Identifier Adapter improves identity consistency and enhances the framework's extensibility, enabling three-characters image animation.}
   \label{ablation1}
\vspace{-0.2cm}
\end{figure*}

\begin{figure}[t]
  \centering
   \includegraphics[width=1\linewidth]{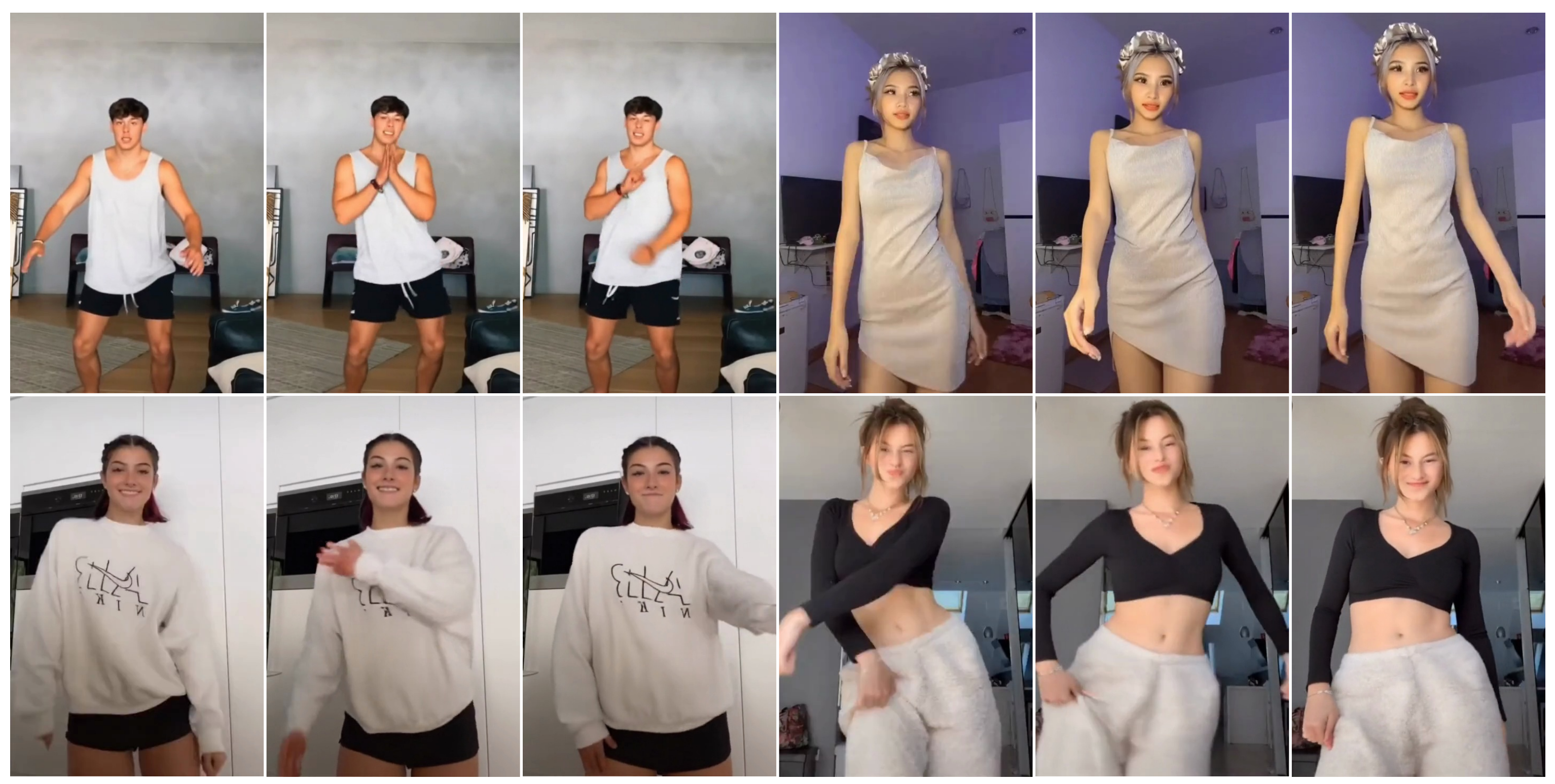}
   \caption{Our framework remains compatible with single-person image animation, even though the introduction increasing training complexity.}
   \label{single}
\vspace{-0.5cm}
\end{figure}

\noindent\textbf{Identifier Assigner \& Identifier Adapter.} 
We evaluate the effectiveness of the mask-driven design. As shown in \cref{ablation1}, the results are obtained with and without the mask-driven design after 40 epochs of training on the Swing Dance dataset \cite{maluleke2024synergy}. The first example, sampled from the Swing Dance dataset, shows that even after sufficient training, the highlighted character in the generated video struggles to maintain temporal consistency. In contrast, our method produces more stable and coherent results.  
The second example is sampled from our collected unseen videos. It indicates that the mask-driven design facilitates the model's capacity to discern three individuals in the generated video, despite being trained exclusively on two-person dataset. This indicates that our design improves both identity consistency and generalizability of our model to unseen multi-person scenarios.

In addition, we use several examples from the TiKToK  \cite{Jafarian_2021_CVPR_TikTok} dataset (which is not used in our model training) to evaluate the compatibility of our model with single-person video generation.
As shown in \cref{single}, even though the introduction of the mask-driven design increases the training complexity, our method still demonstrates strong performance in motion control and identity consistency for single-person videos. 
\begin{table}[h]
    \centering
    \caption{\textbf{Quantitative results on TikTok dataset.} Our approach maintains comparable performance to other models which target at single character animation. }
    \setlength{\tabcolsep}{1.1pt}
    \resizebox{1\linewidth}{!}{
    \begin{tabular}{l|cccc|cc}
        \hline
\multirow{1}{*}{\textbf{Method}} &
\multirow{1}{*}{\textbf{PSNR}$\uparrow$} &
\multirow{1}{*}{\textbf{SSIM}$\uparrow$} &
\multirow{1}{*}{\textbf{L1(E-3)}$\downarrow$} &
\multirow{1}{*}{\textbf{LPIPS}$\downarrow$} &
\multirow{1}{*}{\textbf{FVD}$\downarrow$} &
\multirow{1}{*}{\textbf{FID-VID}$\downarrow$} \\
\hline

DisPose \cite{li2024dispose} & 17.17 & 0.691 & 74.97 & \underline{0.261}  & \underline{615.27} & 64.74\\
        
UniAnimate-DiT \cite{wang2025unianimate} & \underline{17.76} & \underline{0.781} & \underline{68.16} & 0.337  & 649.30 & \underline{50.16}\\

Ours &\textbf{23.68}  &\textbf{0.867} & \textbf{33.65} & \textbf{0.250}  & \textbf{342.48} & \textbf{41.85}\\

    \hline
    \end{tabular}}
    \vspace{-0.2cm}
    \label{tab:single_person}
\end{table}

The quantitative results in \cref{tab:single_person} further show that our approach maintains comparable performance to other state-of-the-art models on single-person generation tasks.

\section{Conclusion}
In conclusion, in this paper, we present an extensible mask-driven framework for multi-character image animation built upon modern Diffusion Transformers. 
By introducing two novel modules-\textbf{Identifier Assigner} and \textbf{Identifier Adapter}-and a dedicated training strategy for identity modeling, our method explicitly models spatial relationships among multiple characters, allowing the framework to extend to varying numbers of participants. 
Furthermore, our training strategy enables the model to generate videos with more characters than those seen during training, significantly improving data efficiency and demonstrating strong practical value. 
Extensive experiments verify that our model, trained only on two-character videos, successfully performs multi-character image animation and achieves superior performance compared with state-of-the-art baselines, while maintaining compatibility with single-person scenarios.

\clearpage
\newpage
\noindent\textbf{Acknowledgment:}
This work is partially supported by the National Natural Science Foundation of China under Grant Number 62476264 and 62406312, the Beijing Natural Science Foundation under Grant Number 4244098, and the Science Foundation of the Chinese Academy of Sciences.
{
    \small
    \bibliographystyle{ieeenat_fullname}
    \bibliography{main}
}


\end{document}